\documentclass[letterpaper]{article} 
\usepackage{aaai25}  
\usepackage{times}  
\usepackage{helvet}  
\usepackage{courier}  
\usepackage[hyphens]{url}  
\usepackage{graphicx} 
\usepackage{natbib}  
\usepackage{caption} 
\usepackage{algorithm}
\usepackage{algorithmic}

\usepackage{newfloat}
\usepackage{listings}
\DeclareCaptionStyle{ruled}{labelfont=normalfont,labelsep=colon,strut=off} 
\floatstyle{ruled}
\newfloat{listing}{tb}{lst}{}
\floatname{listing}{Listing}
\title{C2PD: Continuity-Constrained Pixelwise Deformation for Guided Depth Super-Resolution}
\author{
	Jiahui Kang\textsuperscript{\rm 1},
	Qing Cai\textsuperscript{\rm 1}\thanks{Corresponding author},
	Runqing Tan\textsuperscript{\rm 1},
	Yimei Liu\textsuperscript{\rm 1},
	Zhi Liu\textsuperscript{\rm 2}
}
\affiliations{
	\textsuperscript{\rm 1}Faculty of Computer Science and Technology, Ocean University of China\\
	\textsuperscript{\rm 2}School of Information Science and Engineering, Shandong University\\
 	{\{kangjiahui, trq, liuyimei\}@stu.ouc.edu.cn, cq@ouc.edu.cn, \\  liuzhi@sdu.edu.cn}
}

\usepackage{amsmath}
\usepackage{subfigure}
\usepackage{multirow}
\usepackage{xcolor}
\usepackage{bibentry}

\begin{document}

\maketitle
 
\begin{abstract}
	Guided depth super-resolution (GDSR) has demonstrated impressive performance across a wide range of domains, with numerous methods being proposed. However, existing methods often treat depth maps as images, where shading values are computed discretely, making them struggle to effectively restore the continuity inherent in the depth map. In this paper, we propose a novel approach that maximizes the utilization of spatial characteristics in depth, coupled with human abstract perception of real-world substance, by transforming the GDSR issue into deformation of a roughcast with ideal plasticity, which can be deformed by force like a continuous object. 
	Specifically, we firstly designed a cross-modal operation, Continuity-constrained Asymmetrical Pixelwise Operation (CAPO), which can mimic the process of deforming an isovolumetrically flexible object through external forces. Utilizing CAPO as the fundamental component, we develop the Pixelwise Cross Gradient Deformation (PCGD), which is capable of emulating operations on ideal plastic objects (without volume constraint). Notably, our approach demonstrates state-of-the-art performance across four widely adopted benchmarks for GDSR, with significant advantages in large-scale tasks and generalizability. 
\end{abstract}

\begin{links}
	\link{Code}{https://github.com/amhamster/C2PD}
\end{links}

\section{Introduction}

\begin{figure}[t]
	\centering
	\includegraphics[width=\linewidth]{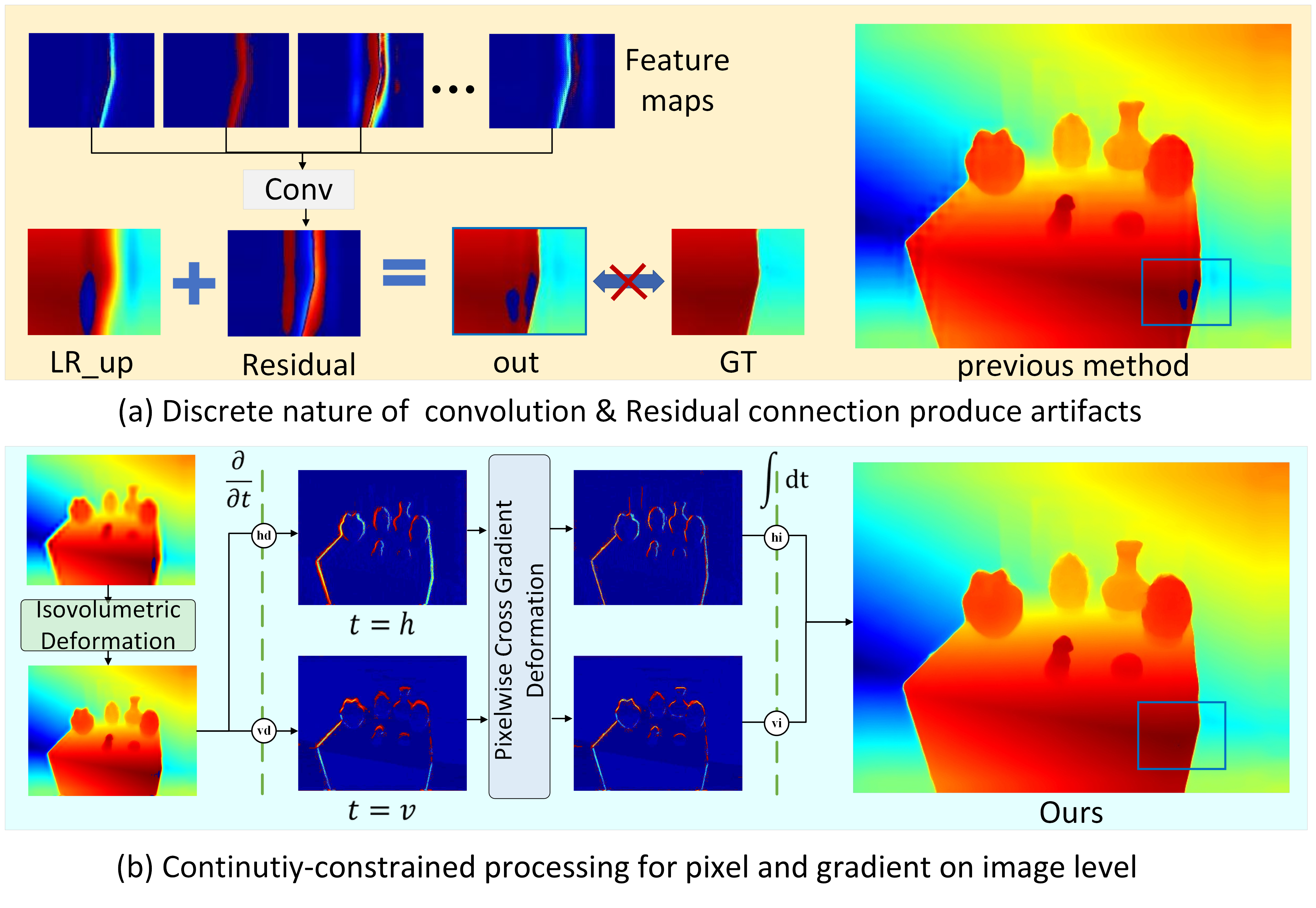}
	\caption{Illustration of some issues of GDSR and our method to address them. (a) After convolution, each point on residual is only obtained by corresponding area of input, which means the relationship of points in output is discrete. Residual connections also increase the difficulty of residual fitting. Those operations treat depth discretely, easy to induce deviation after reconstruction. (b) Our model established a systematic architecture for continuity-constrained deformation with fully integrated operational modules.
	}
\end{figure}

Depth maps are widely utilized in fields such as 3D reconstruction \cite{ref1,ref3}, autonomous driving \cite{ref12}, scene understanding \cite{ref11}, and semantic segmentation \cite{segmentation1,segmentation2,segmentation3}. Technical and cost limitations has prompted the desire to exploit the RGB images to guide super-resolution of depth maps.
Recent research on guided depth super-resolution (GDSR) predominantly centers on deep neural networks, yielding promising outcomes\cite{survey,Guo,PAC,learningGDSR5}. The strategies of computing each point discretely and expanding feature channels, empower fundamental networks like CNN \cite{CNN} and Transformer \cite{Transformer} with outstanding abilities for abstract extraction, which then demonstrate excellent performance in the feature extraction of RGB images \cite{SRCNN,SwinIR,Cai3,Cai1,caihipa,cai4}. Successes in RGB super-resolution have prompted existing methods to process depth at the feature level similarly, merging features of depth and RGB, ultimately reducing channels to obtain the target residual \cite{Residualblock} or depth map.

Although existing methods have achieved impressive performance, as shown in Fig. 1, they often approach depth maps as images where each value is computed discretely, thereby inducing deviation. 
The discrete approaches in previous works led computers to consider depth map as a series of values on a plane, requiring extensive training to learn how to adjust the current value based on other information, lacking the ability to generate holistic spatial awareness akin to humans. In contrast, humans focus on the morphological characteristics rather than specific pixel values. Specifically, when observing low-resolution depth (LR) and ground truth (GT), humans establish associations based on their overall morphological features, unaffected by deviations in object edges or color. This is because our brains can abstract visual information into an object with ideal plasticity, allowing us to imagine the process of LR transforming into GT. Similarly, even when replacing GT with RGB images, we can still envision how LR transforms into GT based on RGB images. Indeed, we can instruct deep learning models to restore LR based on this human awareness. And fortunately, LR, as spatial information already captured, can be directly treated as the initial form of the plasticity model.

Motivated by preceding discussion, we conceptualize the depth as ideal plastic substance, capable of reshaping without volume constraints. 
To integrate continuity constraints into our model, we consider particle interactions and devise the Continuity-constrained Asymmetrical Pixelwise Operation (CAPO) aimed at reshaping the plastic material while maintaining equal volume under external forces. This operation excels in guiding deformation by strictly adhering to source image. 
By exploiting the properties of differential, we propose the Piexlwise Cross Gradient Deformation (PCGD), simulate the complete processing of ideal plastic material and maximize module reusability to ensure model generalization. Moreover, the guidance information utilized by our method differs from specific coloration information in previous methods, emphasizing operational guidance for relative changes (analogous to force information), highlighting abstract information like spatial trends. Consequently, adopting CAPO allows U-net to focus solely on extraction and abstraction, overcoming previous constraints (having to balance the abstract extraction with maintaining cues for reconstruction). In summary, our contributions include:

\begin{enumerate}
	\def\labelenumi{\arabic{enumi}.}
	\item
	A Continuity-constrained Asymmetrical Pixelwise Operation (CAPO) is proposed, which incorporates human perception of continuity into neural networks and can emulate isovolumetric deformation of flexible materials.
	
	\item
	Based on CAPO, a Pixelwise Cross Gradient Deformation (PCGD) is designed, which emulates deformations on objects with ideal plasticity, breaking volume constraints while preserving continuity.
	
	\item
	Extensive experiments demonstrate state-of-the-art performance and exhibit increasingly prominent advantages with scale escalation, suggesting new directions for further breakthroughs in large-scale tasks such as x32.
	
\end{enumerate}
\begin{figure*}[htbp]
	\centering 
	\includegraphics[scale=0.19]{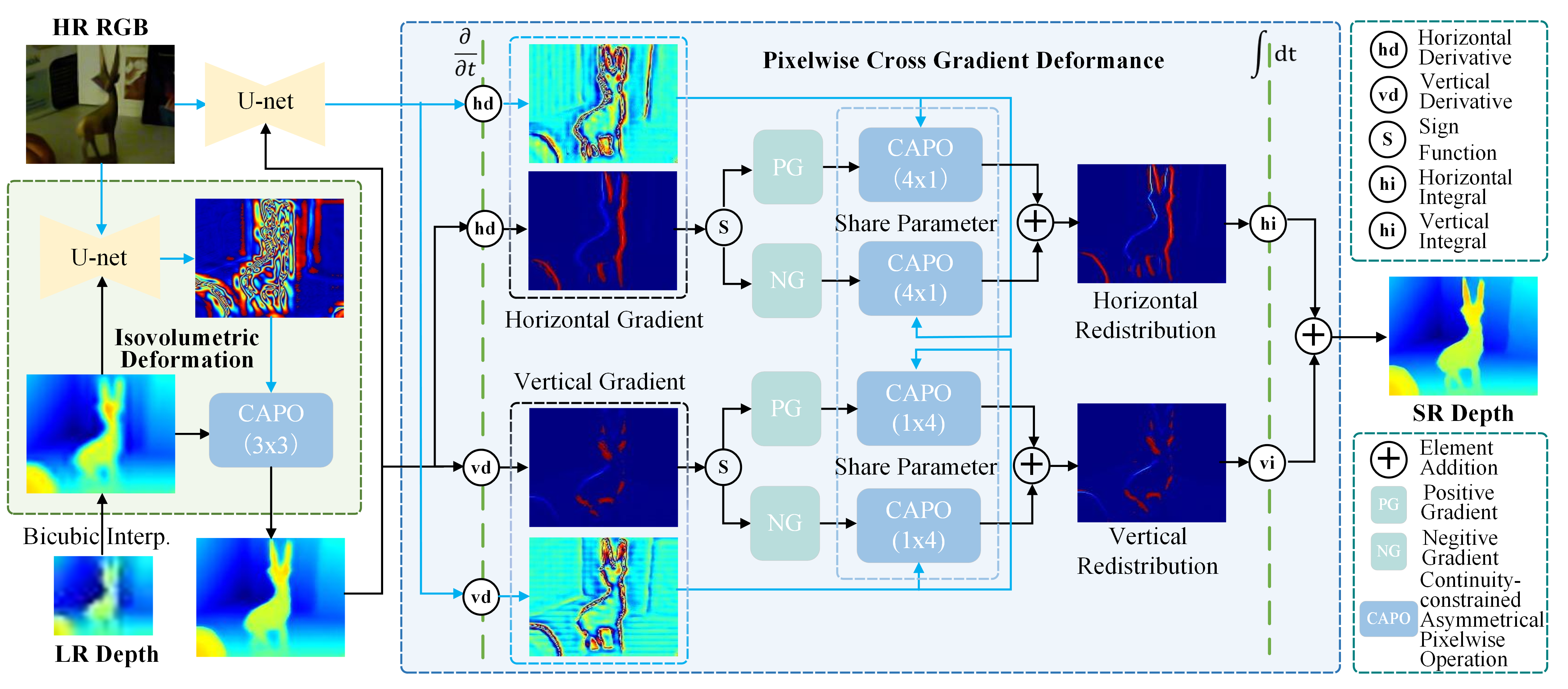}
	\caption{Network architecture of our method. The Isovolumetric Deformation initially uses CAPO to deform the depth to organize pertinent information. Then, depth undergoes further processing through PCGD in the form of gradients.}
\end{figure*}
\section{Relative Work}

\subsection{Deep Learning GDSR Methods}
The potent extraction capability of deep learning has propelled advancements in GDSR \cite{Riegler,Hui,BridgeNet,Ye}.
Early works like DMSG \cite{DMSG}, DG \cite{DG}, and DJF \cite{DJF,DJFR}, initially employed CNNs for feature extraction and direct regression towards the target image. 
\cite{Deng} proposed Deep Coupled ISTA Network based on the Iterative Shrinkage-Thresholding Algorithm (ISTA)\cite{ISTA}, and CUNet \cite{CUNet} for separating common/unique features through sparse coding. 
Furthermore, \cite{DKN} designed a deformable kernel-based filtering network (DKN), adaptively outputting neighborhoods and weights for each pixel. 
\cite{RGBDD} proposed FDSR which exploits octave convolution \cite{oct} to decompose the RGB feature. 
\cite{JIIF} introduced implicit neural representation to learn both the interpolation weights and values (JIIF). \cite{SUFT} introduced a symmetric uncertainty method SUFT, selecting effective RGB information for HR depth recovery while skipping harmful textures. \cite{DCT} developed a discrete cosine network (DCTNet), extracting multi-modal information through a semi-decoupled feature extraction module. Most recently, 
\cite{zhao2023spherical} proposed spherical space feature decomposition (SSDNet) to separate shared and private features.
\cite{DAGF} proposed a multiscale fusion guided filter framework DAGF to reuse intermediate results in the coarse-to-fine process. \cite{DADA} proposed a hybrid approach DADA combining guided anisotropic diffusion with CNNs
for optimization operations on single images. \cite{SGNet} proposed a SGNet designed to capture high-frequency structure from RGB image by emphasizing gradient and frequency domain.

These methods produce pixel values through independent computations, lacking continuity constraints between adjacent regions. Consequently, they are unable to incorporate the rationality of spatial variations as a human would do.

\subsection{Comparison with Existing Methods}
Existing approaches \cite{SGNet,DAGF,DCT} typically involve feature extraction from both depth and RGB modalities, posing challenges in combining the features from RGB and depth to reconstruct continuous information. Moreover, alignment of the two modalities is required before fusion, and deviations in fusion process also affect the results. \cite{align1,align2} In contrast, our method extracts operational information from RGB, providing guidance for depth variations at a global scale, rather than numerically fusing them as similar types. 
Thus, deviations in guidance information hardly misled depth during this operation process. 
Under continuity constraints, depth accepts operational guidance based on its original trend and continuity, meaning that unreasonable textures (may causing by mismatching RGB textures) in the guidance have minimal impact on the depth.

Although DADA \cite{DADA} also attempts to constrain changes in the depth map under guidance of RGB, the process of iterative diffusion is difficult to effectively control, and the resulting images obtained after diffusion cannot be directly used as output. Therefore, DADA ultimately modifies LR\_up by multiplication, rather than directly performing diffuse operations on the source image, and fails to establish a complete and theoretically controlled system for single-image operations. In contrast, our method can directly obtain controlled outputs through CAPO, and effectively operates on gradients with PCGD, ultimately forming a controlled system of continuous deformation without volume limitation.

\section{Method}
Fig. 2 illustrates the architecture of our method.
Firstly, we construct the Continuity-constrained Asymmetrical Pixelwise Operation (CAPO) using neural network.
Subsequently, we derive the formula of the Pixelwise Cross Gradient Deformation (PCGD) and implement its structure.

\begin{figure*}[htbp]
	\centering 
	\includegraphics[scale=0.46]{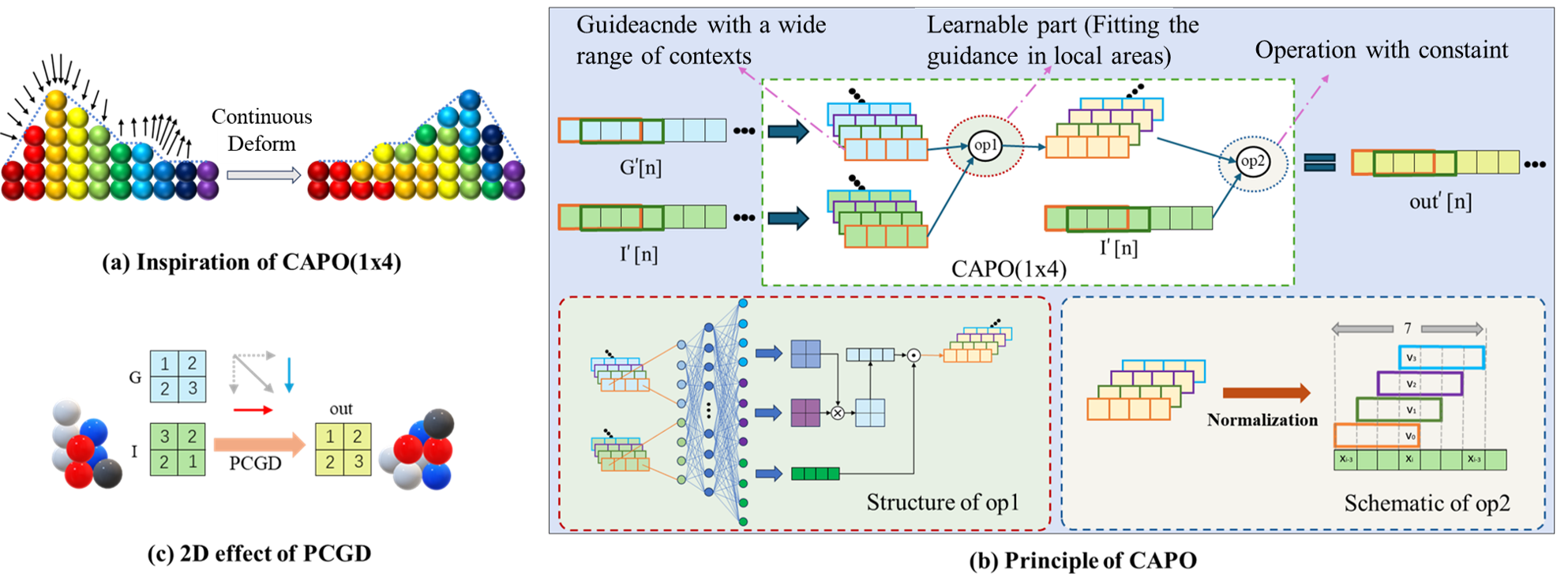} 
	\caption{(a) Considering the interaction among particles, which preferentially convey collectively as a cohesive viscous fluid, the variation at each position is influenced by the combined trends in surrounding regions. (b) The principle of CAPO(1x4) with respect to local information. (c) Controlled with PCGD, each point only needs to focus on its own gradient changes, while the 2D macroscopic effect is regulated by guidance information with large contexts. }
\end{figure*}

\subsection{Continuity-constrained Asymmetrical Pixelwise Operation} 

CAPO is a cross-modal operation designed to model how depth changes under the influence of guidance information, while adhering to continuity constraints. These constraints are based on human cognition of changes in real-world objects, as illustrated in Fig. 3 (a). In this illustration, the value at a point is predominantly derived from its surrounding locations rather than being directly transferred over long distances. In this process, depth and operational information are treated as input, and CAPO can fuse them in a way of guiding isovolumetric flexible object with force field.

Since CAPO(3x3), CAPO(4x1), CAPO(1x4) all share the same idea (just the shape of the receptive field is different), so next we will mainly analyze the CAPO(1x4) operation.
As shown in Fig. 3 (b), the asymmetric structure empowers CAPO to learn the guidance relationship, while the constraints within a region and the effects between regions ensure the continuity of output.
CAPO(1x4) can be seen as ``\(\circ\)'' of \(I_h(x_i) \circ G_h(x_i)\) in formula (9). Here \(I_h(x_i)\) and \(G_h(x_i)\) are two inputs of CAPO, presenting image to be operated and information for guidance respectively.
When considering the horizontal operation, the y coordinate is determined. To simplify the representation, we treat
\(I_h(x_i) \circ G_h(x_i)\) as a one-dimensional operation on the
x-axis (horizontal direction). Among them, \(I_h(x_i)\) is used to represent the variation
of \(I\) along the x direction at position \(x_i\), while \(G_h(x_i)\) is used
to represent the expected change degree at \(x_i\). If \(x_i\) is in the
edge transition area on G, then \(G_h(x_i)\) will be a relatively large
value, which will guide \(I_h\) to assign more values to \(x_i\).
Guided by \(G_h\) crossing modalities, \(I_h\) ought to focus on the relative relationship in local areas of \(G_h\),
rather than directly using \(G_h(x_i)\) as a value for guiding.
Firstly, let us use \(\mathcal {N}^{t}_4\) to represent the four adjacent
positions from x=t-3 to x=t, and use \(D^{t}\) to represent the set of
values on \(I\) and G at these positions, thereby:
\begin{equation}
	D^{t} = \{z|z=I_h(x)\vee G_h(x), x \in \mathcal {N}^{t}_4 \},
\end{equation}
where \(\mathcal {N}^{t}_4 =\{x_{t-3}, x_{t-2},x_{t-1},x_{t}\}\), and ``\(\vee\)'' means ``or'' operation.
To simulate the interaction of particles in a unit area under external force field (op1 in Fig. 3 (c)), we fit \(var_i\) to the target variation at position i within \(\mathcal {N}^{t}_4\), then:
\begin{equation}
	\begin{split}
		Interaction(D^{t}) = & \{var_3(D^{t}), var_2(D^{t}), \\ &var_1(D^{t}),var_0(D^{t})\},
	\end{split}
\end{equation}
where \(var_3\), \(var_2\), \(var_1\), \(var_0\) present the target variations of the four positions \(x_{t-3}\), \(x_{t-2}\),
\(x_{t-1}\), and \(x_{t}\) respectively. 
Here \(var_i\) will decide how many values (particles) to allocate to its own position, based on the relationship between \(x_i\) and the entire \(\mathcal {N}^{t}_4\) area, with the guidance of  \(G_h\).
In order to ensure the principle of volume-conserving constraint, we need to convert \(var_i\) to \(var_j^*\):
\begin{equation}
	var_i^*(D^{t})=var_i(D^{t}) - \frac{1}{4}\sum_{j=0}^3{var_j(D^{t})}.
\end{equation}

Following the principle of particle fluidic propagation, we aim for points closer in distance \(x_i\) to exert a greater influence on \(x_i\) with higher probability. Therefore, we implement op2 in Fig. 3(b) by overlapping regional effects.
Then, the value at position \(x_i\) can be expressed as:
\begin{equation}
	I_h(x_i) \circ  G_h(x_i)  =  I_h(x_i) +  \frac{1}{4}\sum_{j=0}^{3}{var_j^*(D^{i+j})}.
\end{equation}
That is to say, the value at \(x_i\) obtained after the
\(I_h(x_i) \circ G_h(x_i)\) is actually affected by the areas from \(x_{i-3}\) to \(x_{i+3}\), 
with positions closer to \(x_i\)
having a higher probability of making a substantial impact. (op2 in Fig. 3(c))

\begin{figure}[t]
	\centering 
	\includegraphics[width=\linewidth]{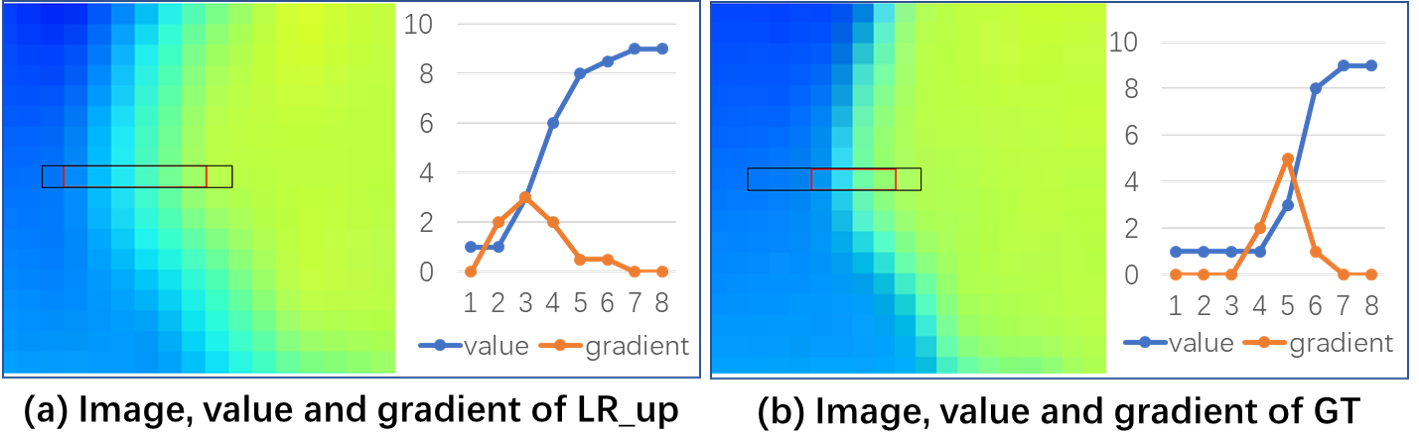}
	\caption{To transfer upsampled LR (LR\_up) to GT, we should move the values to right and shrink the transition area. These changes may be shown as the blue lines, where target values in (a) are all less or equal to the current values in (b), which means we cannot achieve GT by isovolumetric deformation. However, it can be completed by transferring gradient showed as orange line so that the values around position 3 in (a) are passed to position 5, with width shrinking.}
\end{figure}

\subsection{Pixelwise Cross Gradient Deformation}

As shown in Fig. 4, operation on gradient empowers PCGD to genuinely ``move'' edges, rather than merely ``altering'' by blending colors. Thus, we will use this theory as inspiration to develop PCGD.
Let's start by considering real-world scenarios where images are perceived as continuous rather than discrete information. A depth image is essentially the distance from a specific plane to the surface of an object, and surface typically exhibit good continuity. Therefore, we can shift our attention towards the variation trend of depth through derivation, and subsequently adjusting these trends.

\begin{table*}[ht]
	\label{tab:freq}
	\centering
	\resizebox{0.90\textwidth}{!}{
		\begin{tabular}{lccccccccccccccc}
			\hline
			\centering \multirow{2}{*}{Methods} &  \multicolumn{3}{c}{Middlebury} && \multicolumn{3}{c}{Lu} && \multicolumn{3}{c}{NYU\_v2} &&
			\multicolumn{3}{c}{RGBDD} \\
			\cline{2-4}\cline{6-8}\cline{10-12}\cline{14-16}
			& x4 & x8 & x16 && x4 & x8 & x16 && x4 & x8 & x16 && x4 & x8 & x16 \\
			\hline
			\hline
			DKN (IJCV'2021) & 1.23 & 2.12 & 4.24 && 0.96 & 2.16 & 5.11 && 1.62  & 3.26  & 6.51 &&1.30&1.96&3.42\\
			FDKN (IJCV'21) & 1.08 & 2.17  & 4.50  && 0.82 & 2.10 & 5.05 && 1.86  & 3.58  & 6.96 && 1.18&1.91&3.41\\
			FDSR (CVPR'21) & 1.13 & 2.08  & 4.39  && 1.29 & 2.19 & 5.00 && 1.61  & 3.18  & 5.84&&1.18&1.74&3.05\\
			JIIF (MM'21) & 1.09 & 1.82 & 3.31 && 0.85 & 1.73 & 4.16 && 1.37  & 2.76 & 5.27&&1.17&1.79&2.87\\
			DCTNet (CVPR'22) & 1.10 & 2.05 & 4.19 && 0.88 & 1.85 & 4.39 && 1.59  & 3.16 & 5.84&&\underline{1.08}&1.74&3.05\\
			SUFT (MM'22) & 1.07 & 1.75 & 3.18 && 1.10  & 1.74 & 3.92 && \underline{1.12} & 2.51 & 4.86 &&1.10&1.69&2.71\\
			SSDNet (ICCV'23) & \underline{1.02} & 1.91 & 4.02 && \underline{0.80}  & 1.82 & 4.77 && 1.60 & 3.14 & 5.86 &&\textbf{1.04}&1.72&2.92\\
			DAGF (TNNLS'23) & 1.15 & 1.80  & 3.70  && 0.83 & 1.93  & 4.80 && 1.36  & 2.87 & 6.06 &&1.18&1.82&2.91\\
			DADA (CVPR'23) & 1.20 & 2.03 & 4.18 && 0.96 & 1.87 & 4.01 && 1.54 & 2.74 & 4.80&&1.20&1.83&2.80\\
			SGNet (AAAI'24) & 1.15 & \underline{1.64}  & \underline{2.95}  && 1.03 & \underline{1.61} & \underline{3.55} && \textbf{1.10}  & \underline{2.44} & \underline{4.77} &&
			1.10  & \underline{1.64} & \underline{2.55} \\
			
			Ours & \textbf{{0.99}} & \textbf{{1.57}} & \textbf{{2.80}} && \textbf{{0.78}} & \textbf{{1.53}} & \textbf{{3.11}} && 1.25 & \textbf{{2.36}} & \textbf{{4.48}} &&
			1.09&
			\textbf{{1.63}} & \textbf{{2.41}} \\
			\hline
		\end{tabular}
	}
	\caption{Quantitative comparison (in average RMSE) with existing state-of-the-art methods on four benchmark datasets. The best and the second-best values are highlighted by bold and underline, respectively.}
\end{table*}

For a depth map D, since the depth information of each position can be
uniquely determined by the abscissa x and ordinate y, we can represent
the real depth information corresponding to the image through the binary
function \(h (x, y)\), thus the depth map is viewed as a function image of
\(h(x,y)\):
\setlength{\abovedisplayskip}{3pt}
\begin{equation}
	\begin{split}
		D = h(x,y) & =\frac{1}{2}[\int{\frac{\partial h(x,y)}{\partial x}dx}+\int{\frac{\partial h(x,y)}{\partial y}dy}],
	\end{split}
\end{equation}
\setlength{\abovedisplayskip}{\abovedisplayshortskip}
where \(\frac{\partial h(x,y)}{\partial x}\)
and\( \frac{\partial h(x,y)}{\partial y}\) are gradients of depth
in the x and y directions. At this point
we can transform the image D into what we want by adjusting
\(\frac{\partial h(x,y)}{\partial x}\) and
\(\frac{\partial h(x,y)}{\partial y}\).
Now, let us apply the idea in Fig. 4 to explore the method of restoring a blurred LR\_up image to a clear high-resolution image. Whether its blur caused by interpolation or other factors, the image that gives us a blurry
experience is essentially a combination of valid information and biased or erroneous messages. For the blurred image \(I\), we hope to obtain the guidance image \(G\) to guide \(I\) with the help of a specific operation. In the same
way, we can use functions \(f(x,y)\) and \(g(x,y)\) to represent \(I\) and \(G\)
respectively:
\begin{equation}
	\begin{aligned}
		I = f(x,y), G = g(x,y) .
	\end{aligned}
\end{equation}
In other words, the gradients of G, \(\frac{\partial g(x,y)}{\partial x}\) and \(\frac{\partial g(x,y)}{\partial y}\) , can be utilized through an
operation ``\(\circ\)'' to guide the trend of \(I\), \(\frac{\partial f(x,y)}{\partial x}\) and
\(\frac{\partial f(x,y)}{\partial y}\), thereby directing \(I\) convert to
clear image \(I_{out}\) (In fact, as analysis in Fig. 4, we choose CAPO as ``\(\circ\)'') .
Using formula (5)(6), we can get the \(I_{out}\):
\begin{equation}
	\begin{aligned}
		I_{out} =& \frac{1}{2}[\int{(\frac{\partial f(x,y)}{\partial x} \circ \frac{\partial g(x,y)}{\partial x}) dx} \\
		& + \int{(\frac{\partial f(x,y)}{\partial y} \circ \frac{\partial g(x,y)}{\partial y})dy}].
	\end{aligned} 
\end{equation}
Thus, value of \( I_{out}\) at \( (x_p,y_p)\) can be expressed as:
\begin{equation}
	\begin{split}
		\left.I_{out}\right|_{(x_p,y_p)} 
		=&\frac{1}{2}\int_{0}^{x_p}{[f_x^{'}(x,y_p) \circ  g_x^{'}(x,y_p)]dx} \\
		& + \frac{1}{2}\int_{0}^{y_p}[{f_y^{'}(x_p,y) \circ  g_y^{'}(x_p,y)]dy}.
	\end{split} 
\end{equation}
Unlike continuous images in the real world, image processed in computer is sampled to discrete points, so we need to discretize the above formula.
To simplify the expression, we let \(I_h(x) = f_x^{'}(x,y_p) \), 
\(G_h(x) = g_x^{'}(x,y_p)\), \(I_v(x) = f_y^{'}(x_p,y) \),
\(G_v(y) = g_y^{'}(x_p,y)\). Thus we getting:

\setlength{\abovedisplayskip}{-9pt}
\begin{flushleft}
\begin{equation}
	\begin{split}
		&I_{out}|_{(x_p,y_p)} \\
		&= \!\! \frac{1}{2}\!\sum_{i=0}^{x_p}{[I_h(x_i)\!\circ\! G_h(x_i)]}\!+\!\frac{1}{2}\!\sum_{i=0}^{y_p}{[I_v(y_i)\!\circ\! G_v(y_i)]}\\
		&= \!\! \frac{1}{2} \! \sum_{i=0}^{x_p}{ \! [f \!(x_{i+1},\! y_p) \!\! - \!\! f \!(x_i, \! y_p)] \! \circ \! [g (x_{i+1},\! y_p) \!\! - \!\! g (x_i, y_p)]} \\
		&+ \!\! \frac{1}{2} \! \sum_{i=0}^{y_p}{ \! [f \!( x_p,\! y_{i+1}) \!\! - \!\! f \!(x_p,\! y_i)] \! \circ \! [g (x_p,\! y_{i+1}) \!\! - \!\! g(x_p,\! y_i)]}.
	\end{split}
\end{equation}
\end{flushleft}
\setlength{\abovedisplayskip}{\abovedisplayshortskip}

The equation (9) represents our mathematical model for deforming the ideal plastic object (depth map). Considering gap between modalities, we use the absolute gradient of guidance information. As shown in Fig. 2, to leverage the reversible discretization of differential operations, we separate the positive and negative gradients, apply the same operation in parallel, finally restore the directional information and add them together to obtain the processed gradient. (Since differentiation is a reversible operation and the information after differentiation is discretized along the differential direction, separately processing the gradients along each direction will not disrupt the continuity after integration. )

As Fig. 3 (c) shows, although each CAPO in PCGD just focus on gradient in one direction, PCGD leverages the context to provide reasonable changes in both horizontal and vertical gradients for each point. The task of CAPO here can be compared to the point-wise multiplication when using an attention map on an image, which can leverage the local information within
the attention map. 
To ensure that improvements stem from our original architecture and operations, we employ U-net \cite{U-Net} in the same manner as DADA (with the exception that our output channels are set to 32, while DADA uses 64) \cite{DADA, ResNet-50, ImageNet}. 

\begin{figure*}[ht]
	\centering 
	\includegraphics[scale=0.29]{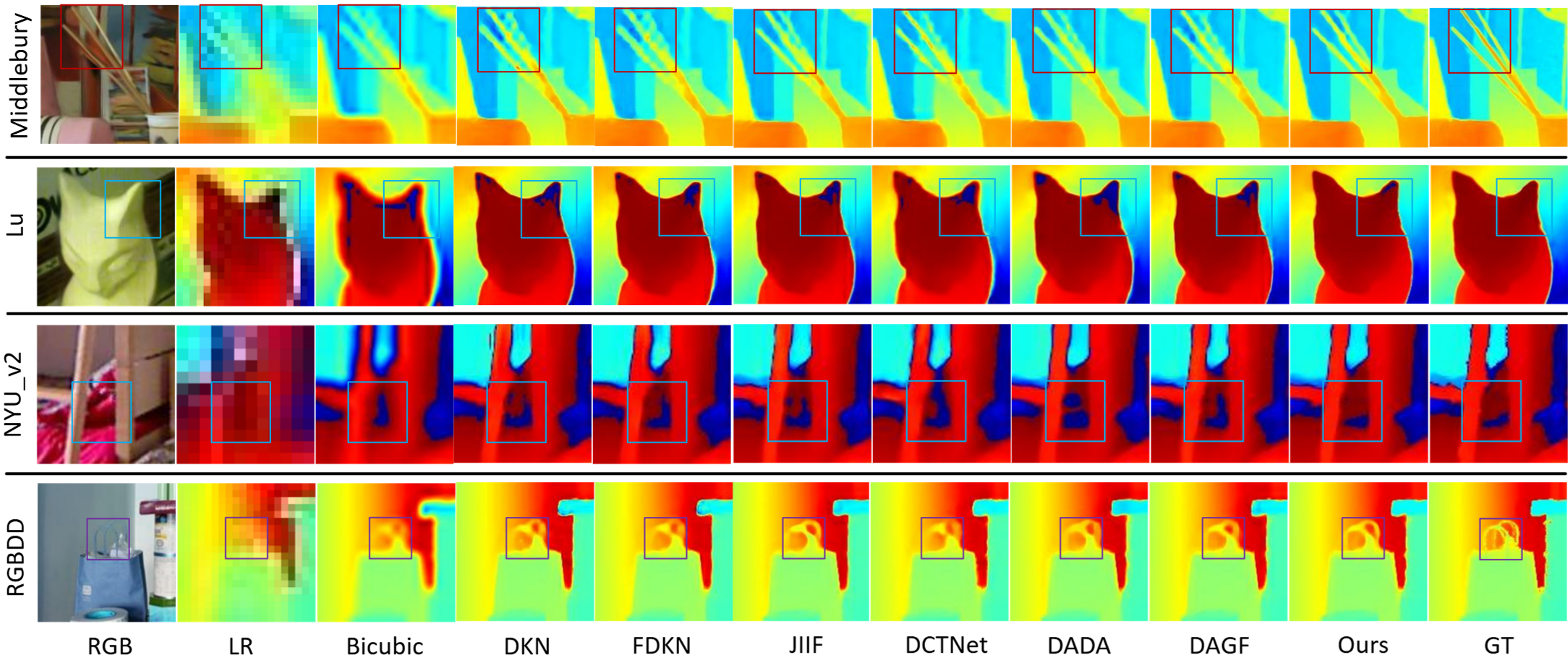}
	\caption{Qualitative comparison of x8 scale factor on four benchmark datasets. }
\end{figure*}

\subsection{Loss Function}
We choose \(L_1\) loss as the training objective, with \(D_{gt}\) and \(D_{out}\) presenting GT and the predicted depth respectively: 
\begin{equation}
	\mathcal{L} = \left\| D_{gt} - D_{out} \right\|_1.
\end{equation}

\section{Experiment}

\subsection{Experimental Settings}
\subsubsection{Datasets and Evaluation Metrics.}
We conduct experiments on NYU\_v2 \cite{NYUv2}, Middlebury \cite{Middlebury1,Middlebury2}, Lu \cite{Lu}, and RGBDD \cite{RGBDD}. Consistent with prior studies  \cite{DKN,RGBDD,DCT,DAGF,SGNet}, we utilize the first 1000 RGB-D pairs from the NYU-v2 dataset for training, with the remaining 449 pairs reserved for validation. Furthermore, the same pretrained model trained on NYUv2 is evaluated on Middlebury (30 pairs), Lu (6 pairs), and RGBDD (405 pairs) datasets. For the synthetic scenes, low-resolution depth inputs are generated by bicubic\cite{bicubic} downsampling of the high-resolution GT. In line with prior approaches \cite{DKN,DAGF,SGNet}, we employ the root mean square error (RMSE) in centimeters as the evaluation metric. 

\subsubsection{Implementation Details.} 
During the training phase, we randomly crop 256\(\times\)256 image patches from depths and RGB images as inputs. Following \cite{DAGF}, we augment the training data with random flipping and rotation. Adam optimizer is utilized \cite{adam} with \(\beta 1\) = 0.9 and \(\beta 2\) = 0.999, employing an initial learning rate of \(1 \times 10^{- 4}\). 
The model is implemented using PyTorch \cite{pytorch} and trained on one RTX 3090ti GPU. Training typically requires two days for the NYU\_v2 dataset.

\subsection{Comparison with SOTAs}
To validate the performance of our method, we compare our method with SOTA methods on ×4, ×8 and ×16 GDSR, including DKN \cite{DKN}, FDKN\cite{DKN}, FDSR \cite{RGBDD}, JIIF\cite{JIIF}, DCTNet \cite{DCT}, SUFT \cite{SUFT}, SSDNet \cite{zhao2023spherical}, DADA \cite{DADA}, DAGF \cite{DAGF}, and SGNet \cite{SGNet}.

\subsubsection{Quantitative Comparison. }
Tab. 1 demonstrates that C2PD achieves state-of-the-art performance on the Middlebury, Lu, NYU\_v2, and RGBDD datasets. Despite certain results for x4 task are suboptimal (because our method does not use a substantial number of parameters to reconstruct dense effective features), in the x16 task, our method outperforms the second best by 0.15cm (Middlebury), 0.44cm (Lu), 0.29cm (NYU\_v2), and 0.14cm (RGBDD) on RMSE (More metrics can be found in supplementary material).

\subsubsection{Qualitative Comparison.}
Fig. 5 displays the visual comparison between our method and others on x8 tasks (Comparisons for more scale factors are provided in supplementary material). It is evident that our method excels not only in edge details but also in preserving the accuracy of internal information. 
This success is attributed to the capability of PCGD to fully leverage global spatial information.

\begin{figure}[h]
	\centering 
	\includegraphics[scale=0.22]{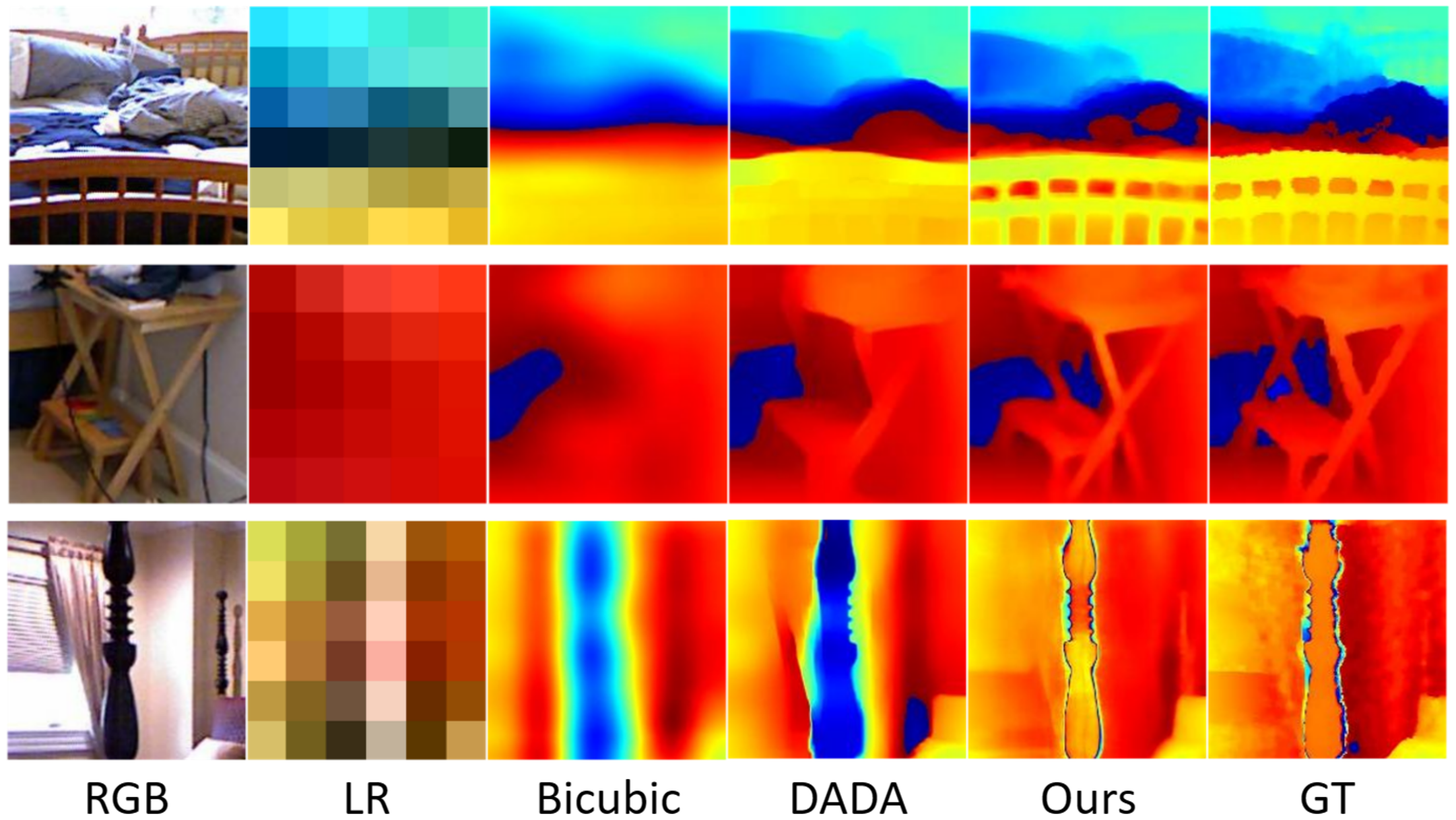}
	\caption{Visual comparison between DADA and our method on NYU\_v2 dataset, with scaling factors of ×32. }
\end{figure}

\begin{table}[h]
	\label{tab:freq}
	\centering 
	\resizebox{0.43\textwidth}{!}{
		\begin{tabular}{ccccc}
			\hline
			Methods&Middlebury&Lu&NYU\_v2&RGBDD\\
			\hline
			DADA & 6.94 & 8.47 & 10.81 & 4.96\\
			Ours & \textbf{5.68} & \textbf{6.48} & \textbf{9.03} & \textbf{4.11} \\
			\hline
		\end{tabular}
	}
	\caption{Quantitative comparison with DADA on x32 scale factor. The best performance is displayed in bold}
\end{table}

\subsubsection{Comparison on Large Scale.}
Tab. 2 and Fig. 6, respectively, present the quantitative and qualitative comparisons on x32 task with DADA, which is demonstrated the best method at the x32 scale. It is evident that our method not only outperforms DADA at the x32 scale but also maintains more detailed and realistic textures without excessive smoothing observed in DADA.

\subsubsection{Complexity Comparison.}
Tab. 3 shows the 
complexity comparison between our method with SOTA method. 
While earlier methods achieved lightweight architectures, recent methods reaching SOTA performance generally incur
signiﬁcantly higher complexity. Our method attains optimal performance with a comparable computational
cost. 

\setlength{\tabcolsep}{2pt}
\begin{table}[h]
	\centering
	\label{tab:freq}
	\scalebox{0.81}{
		\begin{tabular}{cccccc}
			\hline
			Methods&Params (M) &FLOPs (G)&Memory (G) &Time (ms)&RMSE(cm)\\
			\hline
			DAGF & 2.28  & 4623.9 & 10.77 & 73&2.87\\
			DADA & 31.03  & 212.58  & 4.96  & 975 &2.74\\
			SGNet & 39.25 & 4623.9 & 10.77 & 73&2.44\\
			Ours & 62.05 & 414.75 & 8.97 & 61& 2.36\\
			\hline
		\end{tabular}
	}
	\caption{Complexity comparision on NYU (×8)}
\end{table}

\subsection{Ablation Study}

\subsubsection{Ablation Study for CAPO.}
To validate the necessity of CAPO, we fairly replace it with residual convolution (with same parameters amount). As shown in Fig. 7 (a), convolution is far from achieving our effect, highlighting the critical role of CAPO (The difference between CAPO and convolution is discussed in supplementary material).

\begin{figure}[ht]
	\centering
	\includegraphics[width=\linewidth]{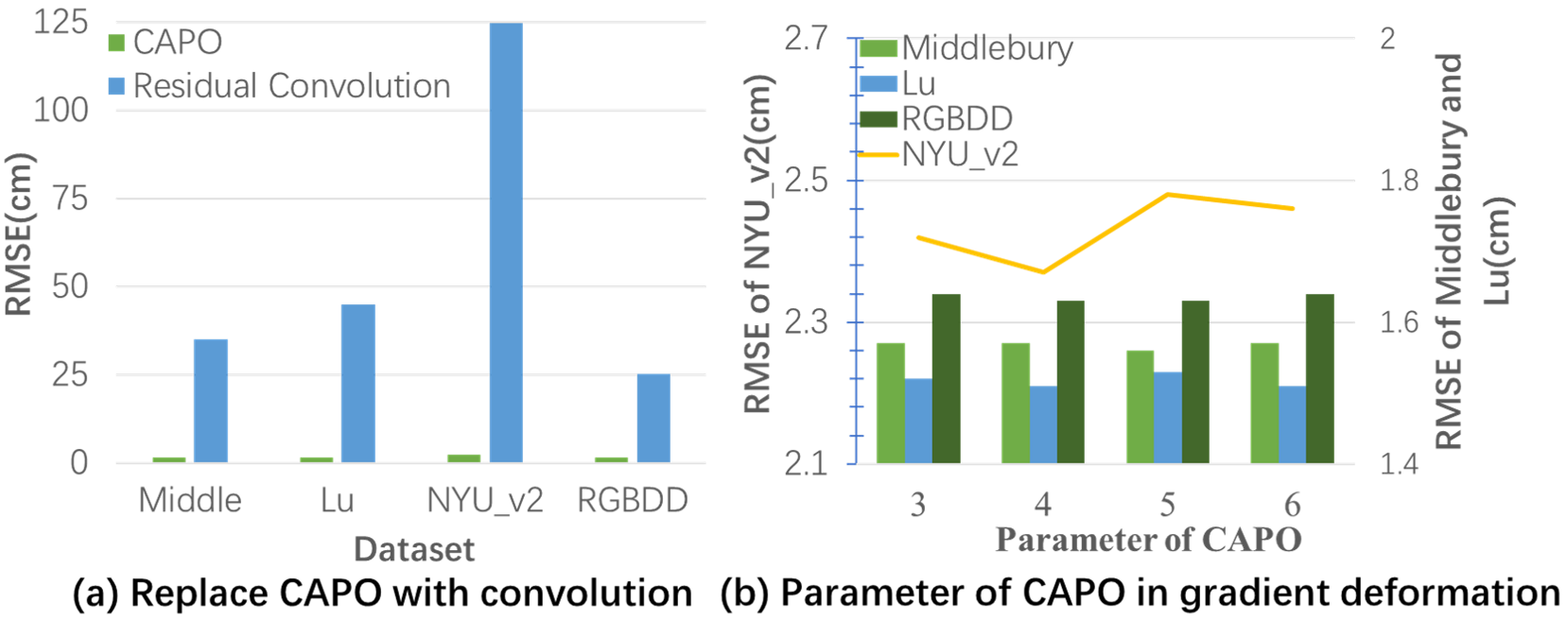}
	\caption{Ablation study for CAPO. (a) Replace CAPO with residual block. (b) Parameter of CAPO in PCGD. }
\end{figure}

Fig. 7 (b) presents the ablation study on parameter of CAPO, which operates on gradients. It is evident that our method achieves optimal performance when the length is set to 4, with stable performance on remaining datasets. (For lengths of 3, 5, and 6, matrix multiplication is replaced by dot products due to computational constraints, potentially contributing to the superior performance at length = 4.)

\subsubsection{Ablation Study for PCGD.}
As shown in Fig. 8 (a), we test four possible structure for PCGD: operating on vertical gradient, operating on horizontal gradient, operating on both direction without sharing parameter, both direction operated in parallel with parameter sharing. When PCGD sharing paramete in parallel, the performance comes to the best. 

\begin{figure}[h]
	\centering
	\includegraphics[width=\linewidth]{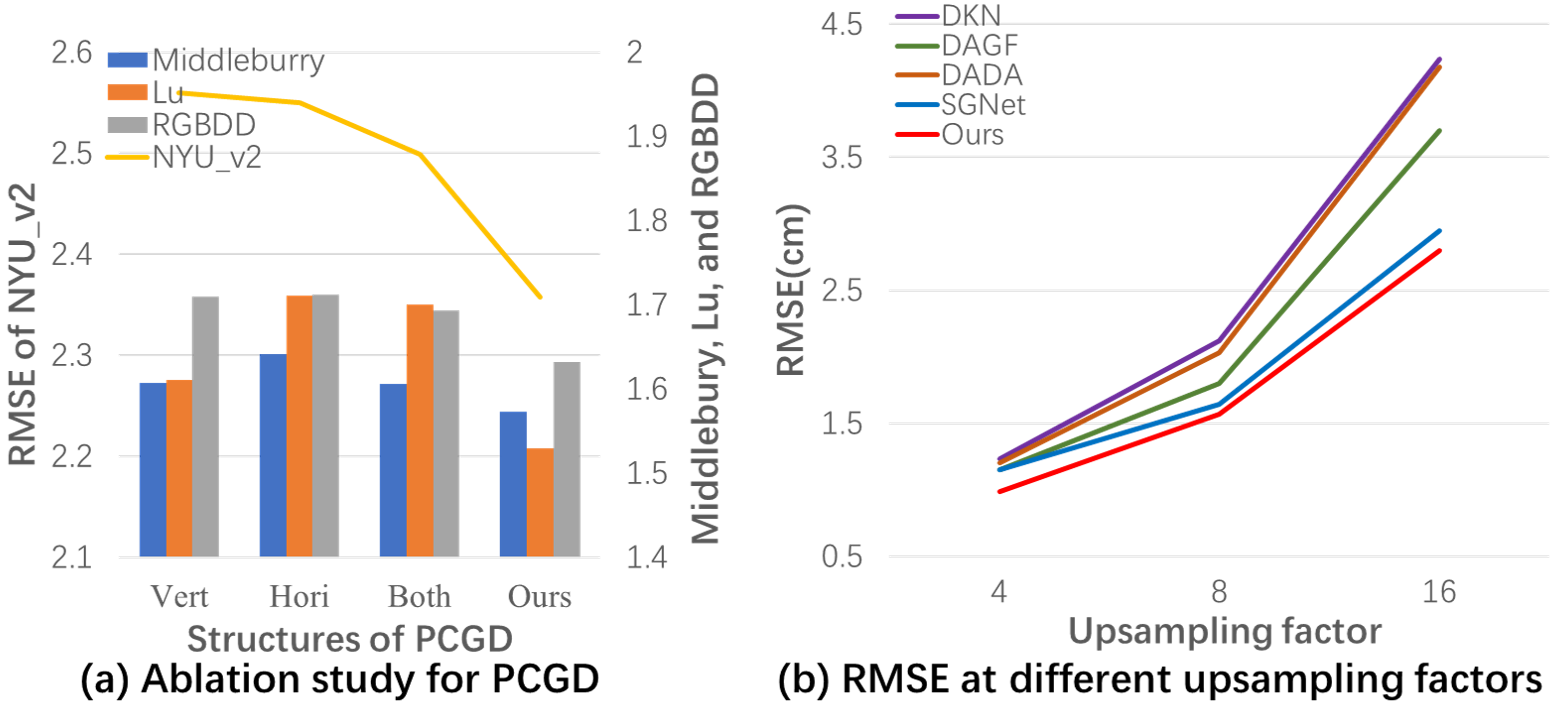}
	\caption{(a) Different structures of PCGD on x8 scale. (b) Comparison at different upsampling factors on Middlebury.}
\end{figure}

\subsection{Further Analysis}

\subsubsection{Analysis for Upsampling Factor.}
As depicted in Fig. 8 (b), our model exhibits a linear degradation trend, indicating that its advantages become more pronounced as the scale increases, confirming that our method can sustain superior performance even with low density of effective information.

\begin{figure}[h]
	\centering 
	\includegraphics[scale=0.32]{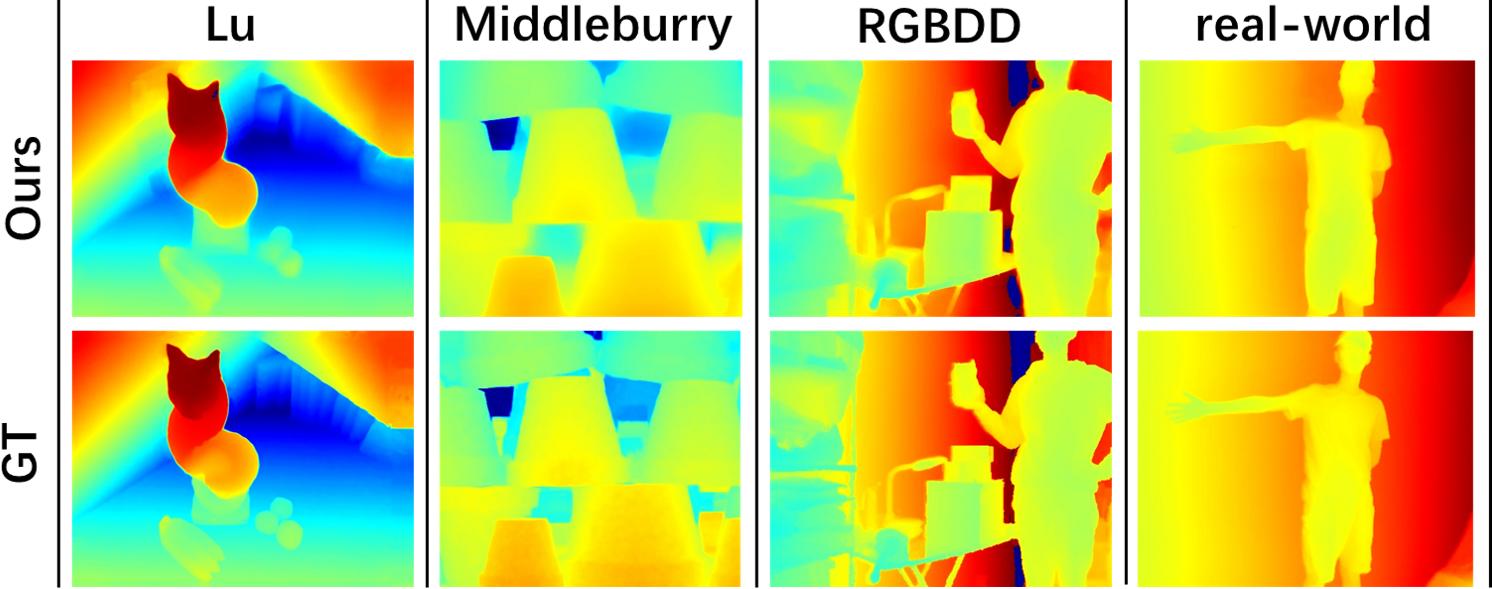}
	\caption{Generalization performance of our x32 model.}
\end{figure}

\subsubsection{Generalization Analysis.} 
Since all methods were trained on the NYU\_v2 , the performance on Middlebury, Lu, and RGBDD reflects the generalization ability. 
Tab. 1 shows our model significantly outperforms the second best on these datasets, indicating strong generalization.

\setlength{\tabcolsep}{3pt}
\begin{table}[h]
	\label{tab:freq}
	\centering
	\resizebox{0.45\textwidth}{!}{
		\begin{tabular}{ccccccc}
			\hline
			Methods&DKN&FDSR&DCTNet&SUFT&SGNet&\textbf{Ours}\\
			\hline
			RMSE &7.38& 7.50 & 7.37 & \underline{7.22} & \underline{7.22}& \textbf{6.68} \\
			\hline
		\end{tabular}
	}
	\caption{Quantitative comparison on real-world RGBDD. }
\end{table}

The robust generalization of our model enhances its adaptability to the intricate scenarios in the real world. Fig. 9 reveals that our trained model (NYU\_v2, x32) exhibits excellent generalization across various datasets, and Tab. 4 presents a quantitative comparison of it with state-of-the-art methods on real-world RGB-D data, demonstrating its significant practical value for real-world applications.

\begin{figure}[h]
	\centering 
	\includegraphics[scale=0.32]{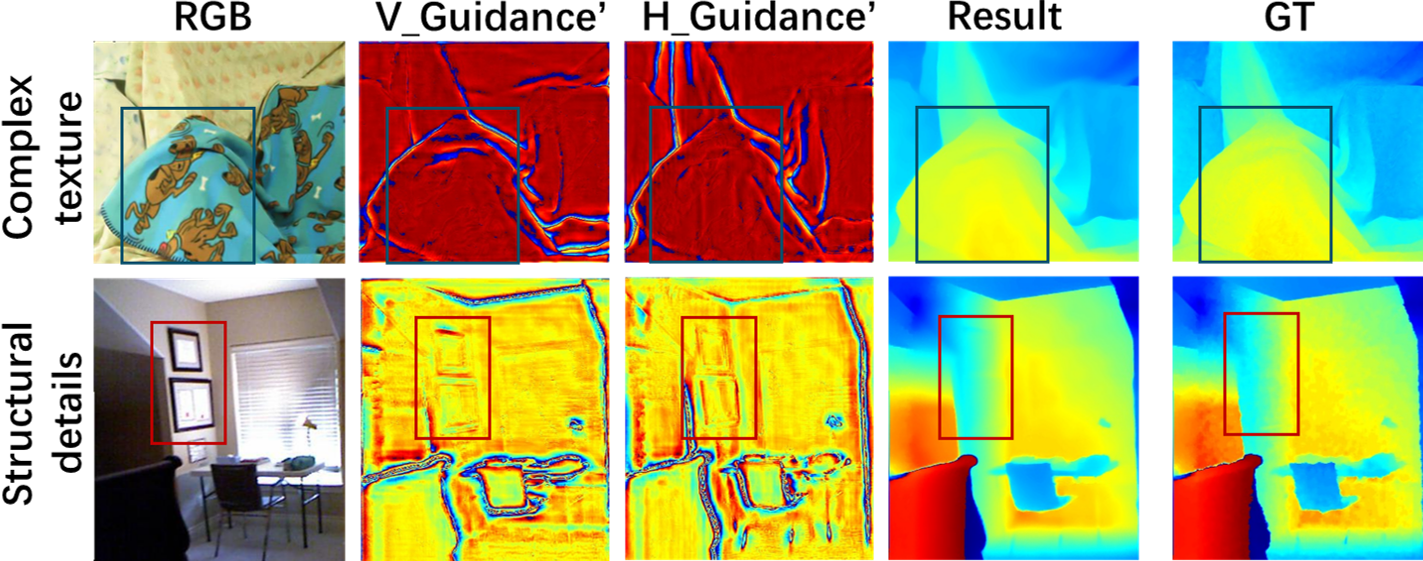}
	\caption{Robustness to RGB textures.}
\end{figure}

\subsubsection{Robustness to RGB.}
Fig. 10 shows the robustness to complex RGB images, which suggest that our approach is not misled by mismatching textures in the guidance. This is because:
(1) Freed from the constraint of ``clear and accurate'', we employ a deeper network to sufficiently highlight and abstract guidance information, thereby weakening and even eliminating complex textures.
(2) As illustrated in Fig. 3 (b), the principle of CAPO is not about merging values but about fitting the deformation pattern of I'[n] using the sequences from G'[n] and I'[n]. 
Even if there are slight traces left by irrelevant structural details, ``op1'' in CAPO can leverages the relationship between the guidance and the depth to weaken erroneous guidance. Additionally, constrained by ``op2'', the deformation of I'[n] is still affected by the values of I'[n] itself, meaning that trivial textures are unlikely to introduce erroneous textures on the smooth depth map.

\section{Conclusion}

In this paper, we propose a novel GDSR method CPGD that integrates human understanding of material properties. By leveraging depth as spatial information, the CAPO operation is proposed, which facilitates continuous constraint deformation on depth, treating it as a flexible object with constant volume. Building upon CAPO, the PCGD is designed, which enable transformations that surpass volume constraints, resulting in a system capable of deforming depth as a substance with ideal plasticity. Our method achieves state-of-the-art performance on four widely used datasets, especially advantageous in large-scale tasks and generalization. This presents a new strategy for further breakthrough in large-scale tasks and real-world application.

\section{Acknowledgements}

This work was supported in part by the National Science Foundation of China under Grant62471448, 62102338; in part by Shandong Provincial Natural Science Foundation under Grant ZR2024YQ004; in part by TaiShan Scholars Youth Expert Program of Shandong Province under Grant No.tsqn202312109.

\bibliographystyle{aaai25}
\bibliography{ref}

\end{document}